\pdfoutput=1
\documentclass[11pt]{article}

\usepackage[hyperref]{acl}
\usepackage{times}
\usepackage{latexsym}

\usepackage{microtype}

\usepackage{times}
\usepackage{latexsym}
\usepackage{amsthm,amsmath,amssymb}
\usepackage{multirow}
\usepackage{array}
\usepackage{mathrsfs}
\usepackage{amsmath}
\usepackage{amsfonts}
\usepackage{graphicx}
\usepackage{booktabs}
\usepackage{algorithm}
\usepackage{algorithmicx}
\usepackage{algpseudocode}

\usepackage{amsfonts}
\usepackage{yhmath}
\usepackage{url}
\usepackage{tikz}
\usepackage{pgfplots}
\usepackage{makecell}
\usepackage{array}
\usepackage{tabularx}
\usepackage{arydshln} 

\usepackage{subcaption}
\usetikzlibrary{shapes, positioning, decorations.pathreplacing, shapes.multipart}

\usepackage{scrextend}

\usepackage{times}
\usepackage{latexsym}

\usepackage[T1]{fontenc}

\usepackage[utf8]{inputenc}

\usepackage{microtype}

\title{An Embarrassingly Easy but Strong Baseline for Nested Named Entity Recognition}

\author{Hang Yan, Yu Sun, Xiaonan Li, 
Xipeng Qiu\thanks{\ \  Corresponding author.}\\
  Shanghai Key Laboratory of Intelligent Information Processing, Fudan University \\
  School of Computer Science, Fudan University \\
  \texttt{\{hyan19,lixn20,xpqiu\}@fudan.edu.cn}\\
  \texttt{yusun21@m.fudan.edu.cn}\\}

\begin{document}
\maketitle
\begin{abstract}
  Named entity recognition (NER) is the task to detect and classify the entity spans in the text. When entity spans overlap between each other, this problem is named as nested NER. Span-based methods have been widely used to tackle the nested NER. Most of these methods will get a  score $n \times n$ matrix, where $n$ means the length of sentence, and each entry corresponds to a span. However, previous work ignores spatial relations in the score matrix. In this paper, we propose using Convolutional Neural Network (CNN) to model these spatial relations in the score matrix. Despite being simple, experiments in three commonly used nested NER datasets show that our model surpasses several recently proposed methods with the same pre-trained encoders. Further analysis shows that using CNN can help the model find more nested entities. Besides, we found that different papers used different sentence tokenizations for the three nested NER datasets, which will influence the comparison. Thus, we release a pre-processing script to facilitate future comparison\footnote{Code is available at \url{https://github.com/yhcc/CNN_Nested_NER}.}.
\end{abstract}

\section{Introduction} 


Named Entity Recognition (NER) is the task to extract entities from raw text. 
It has been a fundamental task in the Natural Language Processing (NLP) field. Previously, this task is mainly solved by the sequence labeling paradigm   
through assigning a label to each token~\cite{DBLP:journals/corr/HuangXY15,DBLP:conf/acl/MaH16,DBLP:journals/corr/abs-1911-04474}. However, this method is not directly applicable to the nested NER scenario, since a token may be included in two or more entities. To overcome this issue, the span-based method which assigns labels to each span was introduced~\cite{DBLP:conf/ecai/EbertsU20,DBLP:conf/acl/LiFMHWL20,DBLP:conf/acl/YuBP20}. 

\citet{DBLP:conf/ecai/EbertsU20} used a pooling method over token representations to get the span representation, and then conducted classification on this span representation. \citet{DBLP:conf/acl/LiFMHWL20} transformed the NER task into a Machine Reading Comprehension form, they used the entity type as the query, and asked the model to select the spans that belong to this entity type. \citet{DBLP:conf/acl/YuBP20} utilized the Biaffine decoder from dependency parsing~\cite{DBLP:conf/iclr/DozatM17} to convert the span classification into classifying the start and end token pairs. However, these work did not take advantage of the spatial correlations between adjacent spans.

\begin{figure}[!t]
  \centering
  \includegraphics[width=\columnwidth]{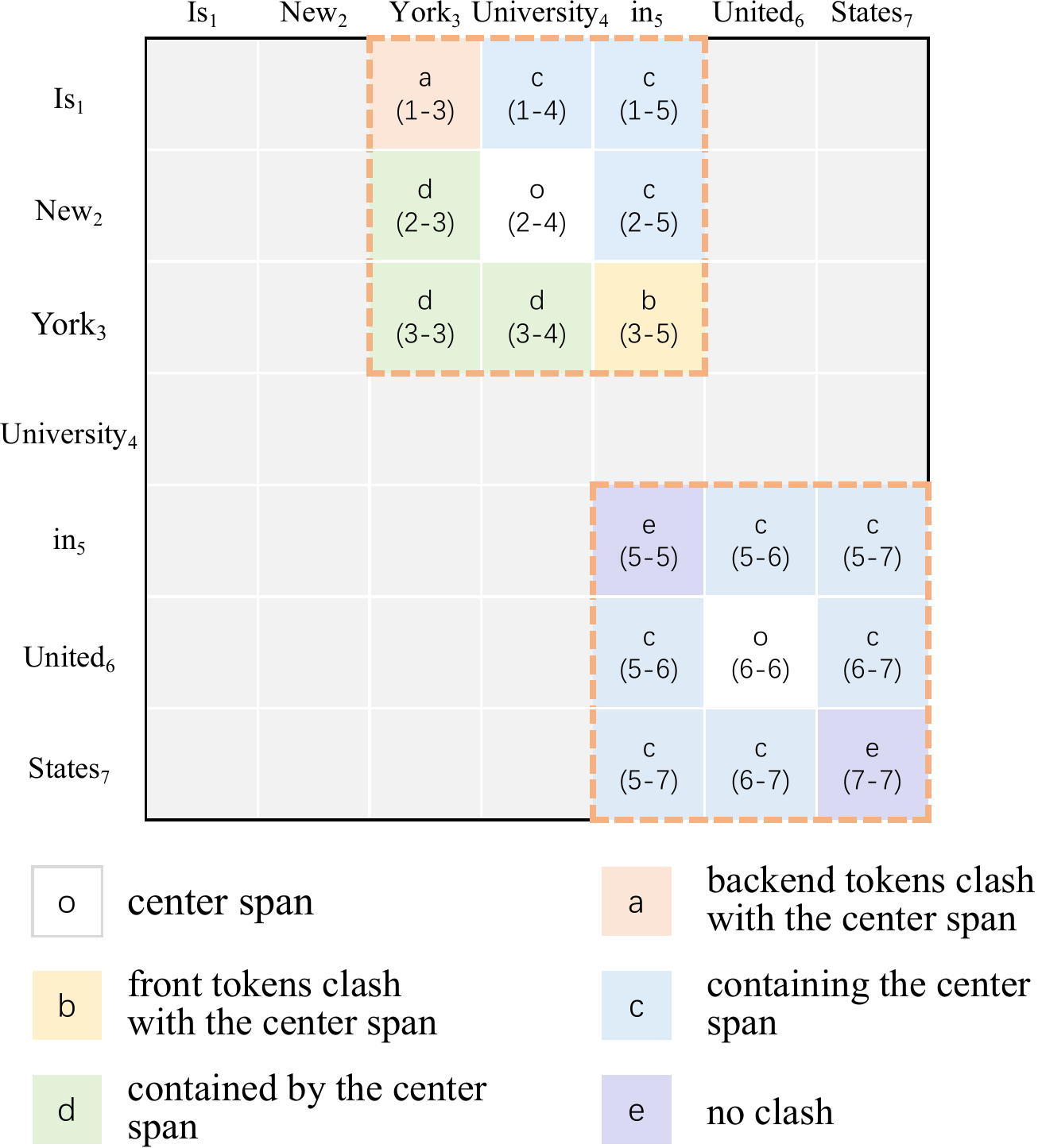}
  \caption{All valid spans of a sentence. We use the start and end tokens to pinpoint a span, for instance, ``(2-4)'' represents ``New York University''. Spans in the two orange dotted squares indicates that the center span can have the special relationship (different relations are depicted in different colors) with its surrounding spans. For example, the span ``New York''~(2-3) is contained by the span ``New York University''~(2-4).  Therefore, the ``(2-3)'' span is annotated as ``d''.} \label{fig:example}
\end{figure}

As depicted in Figure~\ref{fig:example}, the spans surrounding a span have special relationships with the center span. It should be beneficial if we can leverage these spatial correlations. In this paper, we use the Biaffine decoder \cite{DBLP:conf/iclr/DozatM17} to get a 3D feature matrix, where each entry represents one span. After that, we view this feature matrix as an image and utilize Convolutional Neural Network (CNN) to model the local interaction between spans. 

We compare this simple method with recently proposed methods~\cite{DBLP:conf/acl/WanR0022,DBLP:conf/aaai/Li00WZTJL22,DBLP:conf/acl/Zhu022,DBLP:conf/acl/0002THH22}. To make sure our method is strictly comparable to theirs, we asked the authors for their version of data. Although all of them used the same datasets, we found that the statistics, such as the number of sentences and entities, were not the same. This was caused by the usage of distinct sentence tokenization methods, which will influence the performance as shown in our experiments.
To facilitate future comparison, we release a pre-processing script\footnote{\url{https://github.com/yhcc/CNN_Nested_NER/tree/master/preprocess}} for ACE2004, ACE2005 and Genia datasets. 

Our contributions can be summarized as follows.
\begin{itemize}
  \item We find that the adjacent spans have special correlations between each other, and we propose using CNN to model the interaction between them. Despite being very simple, it achieves a considerable performance boost in three widely used nested NER datasets.
  \item We release a pre-processing script for the three nested NER datasets to facilitate direct and fair comparison. 
  \item The way we view the span feature matrix as an image shall shed some light on future exploration of span-based methods for nested NER task.
\end{itemize}

\section{Related Work}
Previously, four kinds of paradigms have been proposed to solve the nested NER task. 

The first one is the sequence labeling framework~\cite{DBLP:conf/acl/StrakovaSH19}, since one token can be contained in more than one entities, the Cartesian product of the entity labels are used. However, the Cartesian labels will suffer from the long-tail issue. 

The second one is to use the hypergraph to efficiently represent spans \cite{DBLP:conf/emnlp/LuR15,DBLP:conf/emnlp/MuisL16,DBLP:conf/naacl/KatiyarC18,DBLP:conf/emnlp/WangL18}. The shortcoming of this method is the complex decoding. 

The third one is the sequence-to-sequence (Seq2Seq) framework \cite{DBLP:conf/nips/SutskeverVL14,DBLP:conf/acl/LewisLGGMLSZ20,DBLP:journals/jmlr/RaffelSRLNMZLL20} to generate the entity sequence. The entity sequence can be the entity pointer sequence \cite{DBLP:conf/acl/YanGDGZQ20,DBLP:conf/aaai/0001JLLRL21} or the entity text sequence \cite{DBLP:conf/acl/0001LDXLHSW22}. Nevertheless, the Seq2Seq method suffers from the time-demanding decoding. 

The fourth one is to conduct span classification. \citet{DBLP:conf/ecai/EbertsU20} proposed to enumerate all possible spans within a sentence, and use a pooling method to get the span representation. While \citet{DBLP:conf/acl/YuBP20} proposed to use the start and end tokens of a span to pinpoint the span, and use the Biaffine decoder to get the scores for each span. The span-based methods are friendly to parallelism and the decoding is easy. Therefore, this formulation has been widely adopted \cite{DBLP:conf/acl/WanR0022,DBLP:conf/acl/Zhu022,DBLP:conf/aaai/Li00WZTJL22,DBLP:conf/acl/0002THH22}. However, the relation between neighbor spans was ignored in previous work.


\begin{figure}[!t]
  \centering
  \includegraphics[width=\columnwidth]{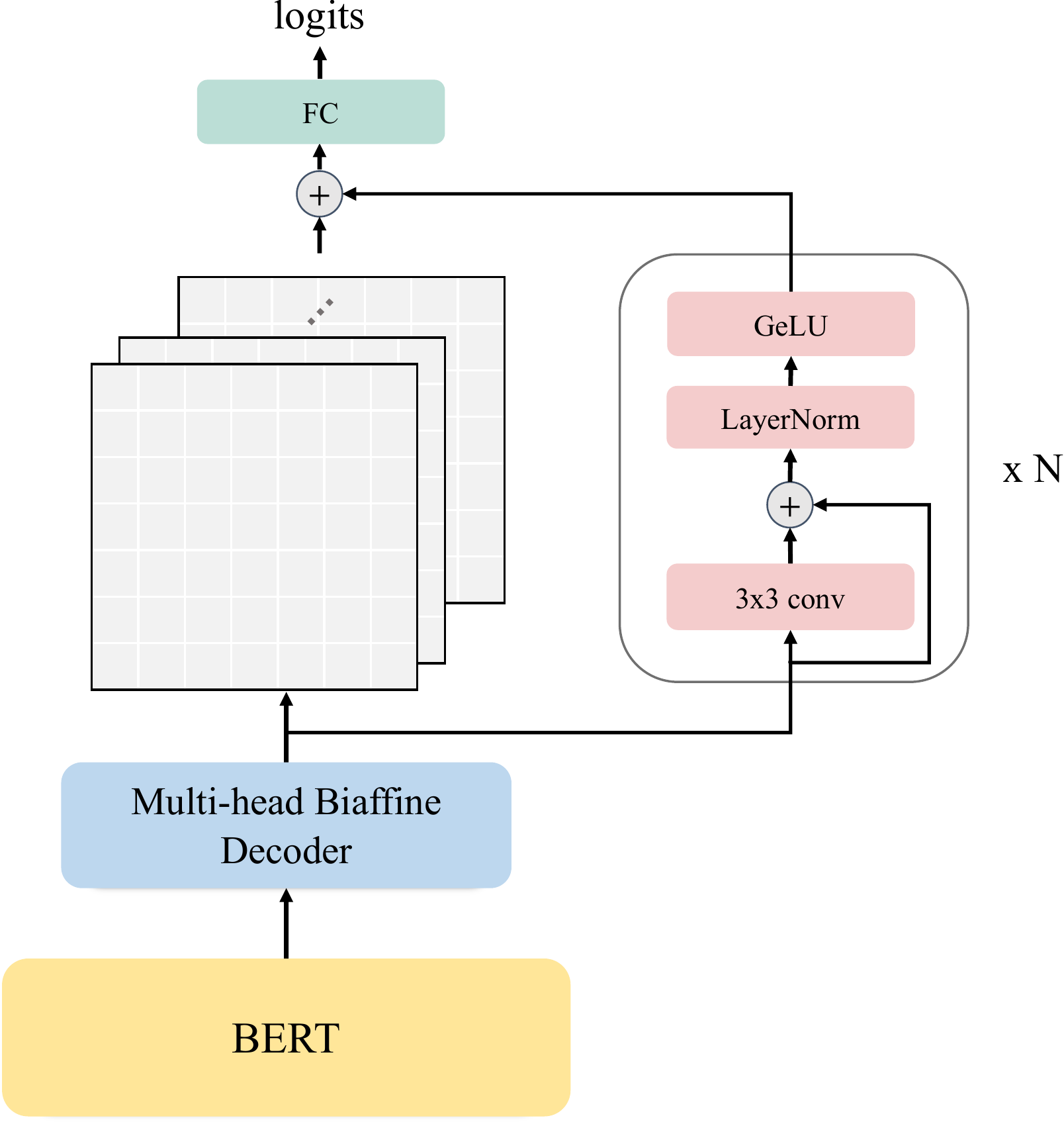}
  \caption{The proposed method in this paper. Use several blocks of CNN to model the spatial correlations between neighbor spans.} \label{fig:model}
\end{figure}

\section{Proposed Method}
In this section, we first introduce the nested NER task, then describe how to get the feature matrix. After that, we present the CNN module to model the spatial correlation on the feature matrix. A general framework of our proposed method can be viewed in Figure~\ref{fig:model}.

\subsection{Nested NER Task}
Given an input sentence $X=[x_1, x_2, \ldots, x_n]$ with $n$ tokens, 
the nested NER task aims to extract all entities in $X$. Each entity can be expressed as a tuple $(s_i, e_i, t_i)$. $s_i, e_i$ are the start, end index of the entity. $t_i \in \{1, \ldots, |T|\}$ is its entity type and $|T|$ is the number of entity types. 
As the task name suggests, the entities may overlap with each other, but different entities are not allowed to have crossing boundaries. For a sentence with $n$ tokens, there are $n(n+1)/2$ valid spans. 

\subsection{Span-based Method for Nested NER}
We follow \citet{DBLP:conf/acl/YuBP20} to formulate this task into a span classification task. Namely, for each valid span, the model assigns an entity label to it. The method first uses an encoder to encode the input sentence as follows:
\begin{align}
  \mathbf{H} = \mathrm{Encoder}(X), \nonumber
\end{align}
where $\mathbf{H} \in \mathcal{R}^{n \times d}$, and $d$ is the hidden size. Various pre-trained models, such as BERT \cite{DBLP:conf/naacl/DevlinCLT19}, are usually used as the encoder. For the word tokenized into several pieces, we use max-pooling to aggregate from its pieces' hidden states.

After getting the contextualized embedding of tokens, previous work usually would concatenate it with the static word embedding and the character embedding, and then send this combined embedding into a BiLSTM layer \cite{DBLP:conf/acl/YuBP20,DBLP:conf/acl/WanR0022,DBLP:conf/acl/0002THH22}. To make the model less cluttered, we neither use more embeddings, nor the BiLSTM layer. 

Next, we use a multi-head Biaffine decoder~\cite{DBLP:conf/iclr/DozatM17,DBLP:conf/nips/VaswaniSPUJGKP17} to get the score matrix as follows:
\begin{align}
  \mathbf{H}_s & = \mathrm{LeakyReLU}(\mathbf{H}W_s), \nonumber \\
  \mathbf{H}_e & = \mathrm{LeakyReLU}(\mathbf{H}W_e),  \nonumber \\
  \mathbf{R} & = \mathrm{MHBiaffine}(\mathbf{H}_s, \mathbf{H}_e) \nonumber
\end{align}
where $W_s,W_e \in \mathcal{R}^{d \times h}$, $h$ is the hidden size, $\mathrm{MHBiaffine}(\cdot, \cdot)$ is the multi-head Biaffine decoder\footnote{The detailed description is in the Appendix.}, and $\mathbf{R} \in \mathcal{R}^{n \times n \times r }$, $r$ is the feature size. Each cell $(i, j)$ in the $\mathbf{R}$ can be seen as the feature vector $\mathbf{v} \in \mathcal{R}^r$ for the span. And for the lower triangle of $\mathbf{R}$ (where $i>j$), the span contains words from the $j$-th to the $i$-th~(Therefore, one span will have two entries if it is off-diagonal).

\subsection{CNN on Score Matrix}
As shown in Figure \ref{fig:example}, the cell has relations with cells around. Therefore, we propose using CNN to model these interactions. We repeat the following CNN block several times in our model:
\begin{align}
  \mathbf{R}' & = \mathrm{Conv2d}(\mathbf{R}), \nonumber \\
  \mathbf{R}'' & = \mathrm{GeLU}(\mathrm{LayerNorm}(\mathbf{R}' + \mathbf{R})), \nonumber
\end{align}
where $\mathrm{Conv2d}$, $\mathrm{LayerNorm}$ and $\mathrm{GeLU}$ are the 2D CNN, layer normalization \cite{DBLP:journals/corr/BaKH16} and GeLU activation function \cite{DBLP:journals/corr/HendrycksG16}. The layer normalization is conducted in the feature dimension. A noticeable fact here is that since the number of tokens $n$ in sentences varies, their $\mathbf{R}$s are of different shapes. To make sure results are the same when $\mathbf{R}$ is processed in batch, the 2D CNN has no bias term, and all the paddings in $\mathbf{R}$ are filled with 0. 

After passing through several CNN blocks, the $\mathbf{R}''$ will be further processed by another 2D CNN module. 

\subsection{The Output}
We use a perceptron to get the prediction logits as follows: \footnote{We did not use the Softmax because in the very rare case (such as in the ACE2005 and Genia dataset), one span can have more than one entity tag.}
\begin{align}
  P = \mathrm{Sigmoid}(W_o(\mathbf{R}+\mathbf{R}'') + b), \nonumber
\end{align}
where $W_o \in \mathcal{R}^{|T|\times r}$, $b \in \mathcal{R}^{|T|}$, $P \in \mathcal{R}^{n \times n \times |T|}$. And then, we use the binary cross entropy to calculate the loss as 
\begin{align}
 \mathcal{L}_{BCE} = - \sum_{0 \le i,j<n}y_{ij}\mathrm{log}(P_{ij}), \nonumber
\end{align}
unlike previous works that only use the upper triangle part to get the loss \cite{DBLP:conf/acl/YuBP20,DBLP:conf/acl/Zhu022},  we use both upper and lower triangles to calculate the loss. The reason is that in order to conduct batch computation, we cannot solely compute the upper triangle part. Since the lower triangle part has been computed, we also use them for the output. The tag for the score matrix is symmetric, namely, the tag in the $(i,j)$-th entry is the same as in the $(j, i)$-th. 

\begin{table*}[!ht]
  \setlength{\tabcolsep}{3pt}
  \centering
  \small
  \begin{tabular}{lp{3em}llllll}
  \toprule
                                              & \multirow{2}{*}{\shortstack[l]{\#  Param. \\(Million)}} & \multicolumn{3}{c}{ACE2004}             & \multicolumn{3}{c}{ACE2005}             \\
                                              \cmidrule(l{8pt}r{8pt}){3-5} \cmidrule(l{8pt}r{8pt}){6-8}
                                              &                            & \multicolumn{1}{c}{P}           & \multicolumn{1}{c}{R}           & \multicolumn{1}{c}{F1}          & \multicolumn{1}{c}{P}           & \multicolumn{1}{c}{R}           & \multicolumn{1}{c}{F1}          \\
  \midrule                                            
  \textit{Data from \citet{DBLP:conf/aaai/Li00WZTJL22}}                &                            &             &             &             &             &             &             \\
  W2NER \cite{DBLP:conf/aaai/Li00WZTJL22}[BERT-large]              & 355.4                      & 87.33       & 87.71       & 87.52       & 85.03       & 88.62       & 86.79       \\
  Ours[BERT-large]                                        & 345.1                     & 87.82$_{38}$ & 87.40$_{20}$ & \textbf{87.61}$_{18}$ & 86.39$_{61}$ & 87.24$_{34}$ & \textbf{86.82}$_{45}$ \\
  \quad w.o. CNN[BERT-large]                                    & 343.6                      & 86.54$_{48}$ & 87.09$_{41}$ & 86.81$_{21}$ & 84.88$_{26}$ & 86.99$_{33}$ & 85.92$_{27}$ \\
  \midrule  
  \textit{Data from \citet{DBLP:conf/acl/WanR0022}}                                      &                            &             &             &             &             &             &             \\
  SG \cite{DBLP:conf/acl/WanR0022}[BERT-base]   & 112.3                        & 86.70       & 85.93       & 86.31       & 84.37       & 85.87       & 85.11       \\
  Ours[BERT-base]                                        & 110.5                      & 86.85$_{61}$ & 86.45$_{36}$ & \textbf{86.65}$_{22}$ & 84.94$_{49}$ & 85.40$_{27}$ & \textbf{85.16}$_{16}$ \\
  \quad w.o. CNN[BERT-base]                                    & 109.1                      & 85.79$_{46}$ & 85.78$_{12}$ & 85.78$_{22}$ & 82.91$_{21}$ & 84.89$_{23}$ & 83.89$_{16}$ \\
  \midrule  
  \textit{Data from \citet{DBLP:conf/acl/Zhu022}}                                            &                            &             &             &             &             &             &             \\
  BS \cite{DBLP:conf/acl/Zhu022}[RoBERTa-base] & 125.6                      & 88.43       & 87.53       & 87.98       & 86.25       & 88.07       & 87.15       \\
  Ours[RoBERTa-base]                                        & 125.6                      & 87.77$_{27}$ & 88.28$_{36}$ & \textbf{88.03}$_{14}$ & 86.58$_{78}$ & 87.94$_{46}$ & \textbf{87.25}$_{48}$ \\
  \quad w.o. CNN[RoBERTa-base]                                    & 125.2                      & 86.71$_{27}$ & 87.40$_{42}$ & 87.05$_{18}$ & 85.48$_{39}$ & 87.54$_{59}$ & 86.50$_{26}$ \\
  \midrule
  \textit{Data from this work}                                   &                            &             &             &             &             &             &             \\
  W2NER[BERT-large]$\dagger$                                       & 355.4                      &      87.17$_{11}$       &     87.70$_{19}$        &     87.43$_{11}$        &        85.78$_{30}$     &      87.81$_{24}$       &     86.77$_{21}$        \\
  Ours[BERT-large]                                        & 345.1                     &   87.98$_{30}$    &    87.50$_{22}$    &   \textbf{87.74}$_{16}$  &       86.26$_{65}$      &       87.56$_{31}$               &    \textbf{86.91}$_{23}$    \\
  \quad w.o. CNN[BERT-large]                                    & 343.6                      &       86.60$_{68}$    &      86.48$_{36}$    & 86.54$_{19}$    &     84.91$_{34}$       &      87.39$_{26}$         &  86.13$_{30}$  \\
  BS[RoBERTa-base]$\dagger$                          & 125.6                      &    87.32$_{40}$         &       86.84$_{16}$      &       87.08$_{24}$      &     86.58$_{38}$        &      87.84$_{59}$       &   87.20$_{32}$          \\
  Ours[RoBERTa-base]                                        & 125.6                      &    87.33$_{41}$    &      87.29$_{25}$       &        \textbf{87.31}$_{16}$     &           86.70$_{29}$    &       88.16$_{54}$      &      \textbf{87.42}$_{26}$     \\
  \quad w.o. CNN[RoBERTa-base]                                    & 125.2                      &     86.09$_{36}$        &    86.88$_{23}$         &   86.48$_{17}$          &      85.17$_{67}$       &      88.0$_{35}$       &  86.56$_{38}$    \\
  \bottomrule
  \end{tabular}
  \caption{Results for the ACE2004 and ACE2005 datasets. Models in the same block use the same data. The subscript means the standard deviation (e.g 87.73$_{18}$ means 87.73$\pm$0.18). $\dagger$ means our reproducation with their publicly available code. } \label{tab:main}
\end{table*}

When inference, we calculate scores in the upper triangle part as:
\begin{align}
  \hat{P_{ij}} = (P_{ij} + P_{ji})/2, \nonumber
\end{align}
where $i\le j$. Then we only use this upper triangle score to get the final prediction. The decoding process generally follows \citet{DBLP:conf/acl/YuBP20}'s method. We first prune out the non-entity spans (none of its scores is above 0.5), then we sort the remained spans based on their maximum entity score. We pick the spans based on this order, if a span's boundary clashes with selected spans, it is ignored.

\section{Experiment}
\subsection{Experimental Setup}
To verify the effectiveness of our proposed method, we conduct experiments in three widely used nested NER datasets, ACE 2004\footnote{\url{https://catalog.ldc.upenn.edu/LDC2005T09}} \cite{DBLP:conf/lrec/DoddingtonMPRSW04}, ACE 2005\footnote{\url{https://catalog.ldc.upenn.edu/LDC2006T06}} \cite{walker2005ace} and Genia \cite{DBLP:conf/ismb/KimOTT03}. 

Besides, we choose recently published papers as our baselines. To make sure our experiments are strictly comparable to theirs, we asked the authors for their version of data. The data statistics for each paper are listed in the Appendix. For ACE2004 and ACE2005, although all of them used the same document split as suggested~\cite{DBLP:conf/emnlp/LuR15}, they used different sentence tokenizations, resulting in different numbers of sentences and entities. To facilitate future research on nested NER, we release the pre-processing code and fix some tokenization issues to avoid including unannotated text and dropping entities. While for the Genia data, we fixed some annotation conflicts (the same sentence with different entity annotations). We replicate each experiment five times and report its average performance with standard derivation.

\begin{table}[!th]
  \centering
  \small
  \setlength{\tabcolsep}{3pt}
  \begin{tabular}{@{}lllll@{}}
  \toprule
  & \multirow{2}{*}{\shortstack[l]{\#  Param. \\(Million)}} & \multicolumn{3}{c}{Genia} \\
  \cmidrule(l{8pt}r{8pt}){3-5}
           &  & P            & R            & F1           \\ \midrule
           \multicolumn{3}{l}{\textit{Data from \citet{DBLP:conf/aaai/Li00WZTJL22}}}             &              &              \\
  W2NER     & 113.6              & 83.10        & 79.76        & 81.39        \\
  Ours      & 112.6              & 83.18$_{24}$ & 79.70$_{8}$  & \textbf{81.40}$_{11}$ \\
  \quad w.o. CNN & 111.1              & 80.66$_{4}$  & 79.76$_{7}$  & 80.21$_{5}$  \\
  \midrule
  \multicolumn{3}{l}{\textit{Data from \citet{DBLP:conf/acl/WanR0022}}}          &              &              \\
  SG        & 112.7              & 77.92        & 80.74        & 79.30        \\
  Ours      & 112.2              & 81.05$_{48}$ & 77.87$_{65}$ & \textbf{79.42}$_{20}$ \\
  \quad w.o. CNN & 111.1              & 78.60$_{41}$ & 78.35$_{52}$ & 78.47$_{16}$ \\
  \midrule
  \multicolumn{3}{l}{\textit{Data from \citet{DBLP:conf/acl/0002THH22}}}     &              &              \\
  Triaffine & 526.5              & 80.42        & 82.06        & 81.23        \\
  Ours      &     128.42               &      83.37$_{9}$            &    79.43$_{15}$  &    \textbf{81.35}${_8}$      \\
  \quad w.o. CNN   &   111.1          &      80.87$_{23}$        &      79.47$_{23}$        &        80.16$_{16}$      \\ 
  \midrule
  \multicolumn{3}{l}{\textit{Data from this work}}     &              &              \\
  W2NER$\dagger$     & 113.6              &     81.58$_{61}$    &    79.11$_{49}$     &      80.32$_{23}$   \\
  Ours      & 112.6              &   81.52$_{21}$ & 79.17$_{18}$  &  \textbf{80.33}$_{13}$ \\
  \quad w.o. CNN & 111.1              & 78.59$_{28}$  & 79.85$_{14}$  & 79.22$_{12}$  \\
  \bottomrule
  \end{tabular}
  \caption{Experiment results for the Genia Dataset. ``W2NER'', ``SG'' and ``Triaffine'' are from~\cite{DBLP:conf/aaai/Li00WZTJL22},~\cite{DBLP:conf/acl/WanR0022} and~\cite{DBLP:conf/acl/0002THH22}, all models use the BioBERT-base\cite{DBLP:journals/bioinformatics/LeeYKKKSK20}. The subscript means the standard deviation (e.g 81.40$_{11}$ means 81.40$\pm$0.11). $\dagger$ means our reproduction with their publicly available code.}
  \label{tab:genia}
\end{table}

\begin{table}[!th]
  \centering
  \small
  \setlength{\tabcolsep}{3pt}
  \begin{tabular}{@{}lllll@{}}
    \toprule
                         & FEP            & FER            & NEP            & NER            \\
                          \midrule
                          \textit{ACE2004}     \\                    
                Ours     & 86.9$_{0.2}$& 87.3$_{0.5}$  & 88.4$_{0.6}$ & \textbf{88.8}$_{0.9}$ \\
         \quad  w.o. CNN & 86.3$_{0.8}$& 86.8$_{0.3}$ & 89.4$_{0.8}$   & 86.6$_{1.3}$\\
                           \midrule
                           \textit{ACE2005} \\
                Ours      & 86.2$_{0.6}$ & 88.3$_{0.1}$  & 91.4$_{0.5}$ & \textbf{89.0}$_{0.8}$ \\
          \quad  w.o. CNN & 85.2$_{0.7}$ & 87.9$_{0.3}$ & 91.3$_{0.5}$ & 86.2$_{0.8}$ \\
                           \midrule
                           \textit{Genia} \\
                Ours     & 81.7$_{0.2}$ & 79.4$_{0.2}$  & 71.7$_{1.6}$ & \textbf{75.5}$_{1.3}$\\
          \quad  w.o. CNN & 79.0$_{0.3}$ & 80.0$_{0.1}$ & 72.7$_{1.2}$  & 64.8$_{1.0}$  \\
          \bottomrule
  \end{tabular}
  \caption{The precision and recall for flat and nested entities in the test set of three datasets. FEP, FER, NEP and NER are the flat entity precision, flat entity recall, nested entity precision and nested entity recall, respectively. Compared with models without CNN (``w.o. CNN''), the most improved metric is bold. By using CNN, the recall for nested entities improve significantly. The subscript means the standard deviation (e.g 88.8$_{0.9}$ means 88.8$\pm$0.9).}
  \label{tab:why_cnn}
\end{table}

\subsection{Main Results}
Results for ACE2004 and ACE2005 are listed in Table~\ref{tab:main}, and for Genia is listed in Table~\ref{tab:genia}. When using the same data from previous work, our simple CNN model surpasses the baselines with less or similar number of parameters, which proves that using CNN to model the interaction between neighbor spans can be beneificial to the nested NER task. Besides, in the bottom block, we reproduced some baselines in our newly processed data to facilitate future comparison. Comparing the last block (processed by us) and the upper blocks (data from previous work), different tokenizations can indeed influence the performance. Therefore, we appeal for the same tokenization for future comparison.

\subsection{Why CNN Helps} \label{sec:why_cnn_help}
To study why CNN can boost the performance of the nested NER datasets, we split entities into two kinds. One kind is entities that overlap with other entities, and the other kind is entities that do not. The results of FEP, FER, NEP, and NER\footnote{The detailed calculation description of the four metrics locate in the Appendix.} are listed in Table~\ref{tab:why_cnn}. Compared with models without CNN, the NEP of models with CNN improved for 2.2, 2.8 and 10.7 for ACE2004, ACE2005 and Genia respectively. Namely, much of the performance improvement can be ascribed to finding more nested entities. This is expected as the CNN can be more effective for exploiting the neighbor entities when they are nested.

\section{Conclusion}
In this paper, we propose using CNN on the score matrix of span-based NER model. Although this method is very simple, it achieves comparable or better performance than recently proposed methods. Analysis shows exploiting the spatial correlation between neighbor spans through CNN can help model find more nested entities. And experiments show that different tokenizations indeed influence the performance. Therefore, it is necessary to make sure all comparative baselines uses the same tokenization. To facilitate future comparison, we release a new pre-processing script for three nested NER datasets.

\bibliography{anthology,custom}
\bibliographystyle{acl_natbib}

\appendix

\begin{table*}[]
  \centering
  \small
  \setlength{\tabcolsep}{3pt}
  \begin{tabular}{clrrrcrrrrc}
  \toprule
  \multicolumn{1}{l}{}     &       & \multicolumn{4}{c}{Sentence}                                                                                        & \multicolumn{5}{c}{Mention}                                                                            \\ \cmidrule(r){3-6} \cmidrule(r){7-11}
  \multicolumn{1}{l}{}     &       & \multicolumn{1}{l}{\#Train} & \multicolumn{1}{l}{\#Dev} & \multicolumn{1}{l}{\#Test} & \multicolumn{1}{l}{Avg. Len} & \multicolumn{1}{l}{\#Ovlp.} & \multicolumn{1}{l}{\#Train} & \multicolumn{1}{l}{\#Dev} & \multicolumn{1}{l}{\#Test} & \multicolumn{1}{l}{Avg. Len}  \\ \midrule
  \multirow{4}{*}{ACE2004} & W2NER & 6,802 & 813 & 897 & 20.12 & 12,571 & 22,056 & 2,492 & 3,020 & 2.5                         \\
                           & SG    & 6.198                        & 742                       & 809                        & 21.55                        & 12,666                       & 22,195                       & 2,514                      & 3,034                       & 2.51                         \\
                           & BS    & 6,799                        & 829                       & 879                        & 20.43                        & 12,679                       & 22,207                       & 2,511                      & 3,031                       & 2.51                         \\
                           & Ours  & 6,297                        & 742                       & 824                        & 23.52                        & 12,690                       & 22,231                       & 2,514                      & 3,036                       & 2.64                         \\
                           \midrule
  \multirow{4}{*}{ACE2005} & W2NER & 7,606 & 1,002 & 1,089 & 17.77 & 12,179 & 24,366 & 3,188 & 2,989 & 2.26                        \\
                           & SG    & 7,285                        & 968                       & 1,058                       & 18.60                         & 12,316                       & 24,700                       & 3,218                      & 3,029                       & 2.26                         \\
                           & BS    & 7,336                        & 958                       & 1,047                       & 18.90                         & 12,313                       & 24,687                       & 3,217                      & 3,027                       & 2.26                         \\
                           & Ours  & 7,178                        & 960                       & 1,051                       & 20.59                        & 12,405                       & 25,300                       & 3,321                      & 3,099                       & 2.40                          \\
                           \midrule
  \multirow{4}{*}{Genia}   & W2NER & 15,023 & 1,669 & 1,854 & 25.41 & 10,263 & 45,144 & 5,365 & 5,506 & 1.97 \\
                           & SG    & 15,022 & 1,669 & 1,855 & 26.47 & 10,412 & 47,006 & 4,461 & 5,596 & 2.07                       \\
                           & Triaffine    & 16,692 & - & 1,854 & 25.41 & 10,263 & 50,509 & - & 5,506 & 1.97                       \\

                           & Ours & 15,038 & 1,765 & 1,732 & 26.47 & 10,315 & 46,203 & 4,714 & 5,119 & 2.0 \\
  \bottomrule                          
  \end{tabular}
  \caption{The statistics used in each paper. ``W2NER''\protect\footnotemark, ``SG'', ``BS'' and ``Triaffine'' are from \cite{DBLP:conf/aaai/Li00WZTJL22}, \cite{DBLP:conf/acl/WanR0022}, \cite{DBLP:conf/acl/Zhu022} and \cite{DBLP:conf/acl/0002THH22}, respectively. Different papers used different sentence tokenization for ACE2004 and ACE2005, resulting in different numbers of sentences in each split. To facilitate future comparison, we open-sourced a pre-processing script to prepare ACE2004 and ACE2005. Previously, some entities will be dropped because of sentence tokenization, we avoid sentence tokenization within an entity and resulting in more entities. And for Genia, different papers used different train/dev/test splits. Besides, the Genia data has conflicting annotations, we remove these sentences. The data annotated with ``Ours'' is obtained by our pre-processing code.}
  \label{tab:statistics}
\end{table*}
\footnotetext{The number of entites is different from that reported in their paper, because we found some duplicated entities in their data.}


\section{Multi-head Biaffine Decoder}
The input of Multi-head Biaffine decoder is two matrix $\mathbf{H}_s,\mathbf{H}_e \in \mathcal{R}^{n \times h}$, and the output is $\mathbf{R} \in \mathcal{R}^{n \times n \times r}$. The formulation of Multi-head Biaffine decoder is as follows
\begin{align}
  \mathbf{S}_1[i, j] & = (\mathbf{H}_s[i]\oplus \mathbf{H}_e[j] \oplus \mathbf{w}_{i-j})W, \nonumber\\
  \{\mathbf{H}_s^{(k)}\}, \{\mathbf{H}_e^{(k)}\} & = \mathrm{Split}(\mathbf{H}_s), \mathrm{Split}(\mathbf{H}_e), \nonumber \\
  \mathbf{S}_2^{(k)}[i,j] & = \mathbf{H}_s^{(k)}[i]U\mathbf{H}_e^{(k)}[j]^T, \nonumber \\
  \mathbf{S}_2 & = \mathrm{Concat}(\mathbf{S}^{(1)}_2, ..., \mathbf{S}^{(K)}_2), \nonumber \\
  \mathbf{R} & = \mathbf{S}_1 + \mathbf{S}_2, \nonumber
\end{align}
where $\mathbf{H}_s, \mathbf{H}_e \in \mathcal{R}^{n \times h}$, $h$ is the hidden size, $\mathbf{w}_{i-j} \in \mathcal{R}^{c}$ is the span length embedding for length $i-j$, $W \in \mathcal{R}^{(2h+c) \times r}$, $\mathbf{S}_1 \in \mathcal{R}^{n \times n \times r}$, $r$ is the biaffine feature size, $\mathrm{Split}(\cdot)$ equally splits a matrix in the last dimension, thus, $\mathbf{H}_s^{(k)},\mathbf{H}_e^{(k)} \in \mathcal{R}^{n \times h_k}$; $h_k$ is the hidden size for each head, and $U \in \mathcal{R}^{h_k \times r_k \times h_k}$, $\mathbf{S}_2 \in \mathcal{R}^{n \times n \times r }$, and $\mathbf{R} \in \mathcal{R}^{n \times n \times r }$.

We did not use multi-head for $W$, because it does not occupy too much parameters and using multi-head for $W$ harms the performace slightly.

\section{Data}
We list the statistics for each datasets in Table \ref{tab:statistics}. As shown in the table, the number of sentences and even the number of entities are different for each paper. Therefore, it is not fair to directly compare results. For the ACE2004 and ACE2005, we release the pre-processing code to get data from the LDC files. We make sure no entities are dropped because of the sentence tokenization. Thus, the pre-processed ACE2004 and ACE2005 data from this work in Table \ref{tab:statistics} have the most entities. And for Genia, we appeal for the usage of train/dev/test, and we release the data split within the code repo. Moreover, in order to facilitate the document-level NER study, we split the Genia dataset based on documents. Therefore, sentences from train/dev/test splits are from different documents, the document ratio for train/dev/test is 8:1:1. Besides, we found one conflicting document annotation in Genia, we fix this confict. After comparing different versions of Genia, we found the W2NER \cite{DBLP:conf/aaai/Li00WZTJL22} and Triaffine \cite{DBLP:conf/acl/0002THH22} dropped the spans with more than one entity tags (there are 31 such entities). Thus, they have less number of nested entities than us. While SG \cite{DBLP:conf/acl/WanR0022} includes the discontinuous entities, so they have more number of nested entities than us.

\section{Implementation Details}
We used the AadmW optimizer to optimize the model and the transformers package for the pre-trained model~\cite{wolf-etal-2020-transformers}. 
The hyper-parameter range in this paper is listed in Table~\ref{tab:hyper}.

\begin{table}[!h]
  \centering
  \small
  \setlength{\tabcolsep}{3pt}
  \begin{tabular}{@{}llll@{}}
    \toprule
                   & ACE2004        & ACE2005        & Genia \\
                   \midrule
  \# Epoch         & 50             & 50             & 5     \\
  Learning Rate    & 2e-5           & 2e-5           & 7e-6  \\
  Batch size       & 48             & 48             & 8     \\
  \# CNN Blocks    & [2, 3]     & [2, 3]     & 3     \\
  CNN kernel size  & 3              & 3              & 3     \\
  CNN Channel dim. & [120, 200] & [120, 200] & 200   \\
  \# Head          & [1, 5]              & [1, 5]              & 4     \\
  Hidden size $h$  & 200            & 200            & 400   \\
  Warmup factor    & 0.1            & 0.1            & 0.1   \\
  \bottomrule
  \end{tabular}
  \caption{The hyper-parameter in this paper.}
  \label{tab:hyper}
  \end{table}

\begin{table}[!h]
  \centering
  \small
  \begin{tabular}{@{}lccc@{}}
    \toprule
          & \# Ent. & \# Flat Ent. & \# Nested Ent. \\
          \midrule
  ACE2004 &   3,036  &  1,614  &  1,422   \\
  ACE2005 &   3,099  &  1,913  &  1,186            \\
  Genia   &   5,119  &  3,963  &  1,156         \\
  \bottomrule            
  \end{tabular}
  \caption{The flat and nested entity statistics in the test set of each dataset. } \label{tab:why_stat}
\end{table}

\section{FEP FER NEP NER}
We split entities into two kinds based on whether they overlap with other entities, and the statistics for each dataset are listed in Table \ref{tab:why_stat}. When calculating the flat entity precision~(FEP), we first get all flat entities in the prediction and calculate their ratio in the gold. For the flat entity recall~(FER), we get all flat entities in the gold and calculate their ratio in the prediction. And we get the nested entity precision~(NEP) and nested entity recall~(NER) similarly.

\end{document}